\documentclass{easychair}
\usepackage{doc}
\usepackage{float}
\usepackage{graphicx}
\usepackage{amsmath}
\usepackage{subcaption}

\title{Object-sensitive Deep Reinforcement Learning}
\author{
Yuezhang Li
\and
   Katia Sycara
\and
   Rahul Iyer
   }

\institute{
  Carnegie Mellon University,
  Pittsburgh, PA, USA\\
  \email{yuezhanl@andrew.cmu.edu, katia@cs.cmu.edu, rahuli@andrew.cmu.edu}
 }

\authorrunning{Yuezhang, Rahul and Katia}

\titlerunning{Object-sensitive DRL}

\begin{document}

\maketitle

\begin{abstract}
  Deep reinforcement learning has become popular over recent years, showing superiority on different visual-input tasks such as playing Atari games and robot navigation. Although objects are important image elements, few work considers enhancing deep reinforcement learning with object characteristics. In this paper, we propose a novel method that can incorporate object recognition processing to deep reinforcement learning models. This approach can be adapted to any existing deep reinforcement learning frameworks. State-of-the-art results are shown in experiments on Atari games. We also propose a new approach called ``object saliency maps'' to visually explain the actions made by deep reinforcement learning agents. 
\end{abstract}



%
%

\section{Introduction}
\label{sect:introduction}
Deep neural networks have been widely applied in reinforcement learning (RL) algorithms to achieve human-level control in various challenging domains. More specifically, recent work has found outstanding performances of deep reinforcement learning (DRL) models on Atari 2600 games, by using only raw pixels to make decisions \cite{mnih2015humanlevel}. 

The literature on reinforcement learning is vast. Multiple deep RL algorithms have been developed to incorporate both on-policy RL such as Sarsa \cite{sutton1996generalization}, actor-critic methods \cite{barto2004j}, etc. and off-policy RL such as Q-learning using experience replay memory \cite{mnih2015humanlevel} \cite{riedmiller2005neural}. A parallel RL paradigm \cite{DBLP:journals/corr/MnihBMGLHSK16} has also been proposed to reduce the heavy reliance of deep RL algorithms on specialized hardware or distributed architectures. However, while a high proportion of RL applications such as Atari 2600 games contain objects with different gain or penalty (for example, enemy ships and fuel vessel are two different objects in the game ``Riverraid''), most of previous algorithms are designed under the assumption that various game objects are treated equally. 

In this paper, we propose a new \textbf{Object-sensitive Deep Reinforcement Learning (O-DRL)} model that can exploit object characteristics such as presence and positions of game objects in the learning phase. This new model can be adapted to most of existing deep RL frameworks such as DQN \cite{mnih2015humanlevel} and A3C \cite{DBLP:journals/corr/MnihBMGLHSK16}. Our experiments show that our method outperforms the state-of-the-art methods by 1\% - 20\% in various Atari games.

Moreover, current deep RL models are not explainable, i.e., they cannot produce human understandable explanations. When a deep RL agent comes up with an action, people cannot understand why it picks the action. Therefore, there is a clear need for deep RL agents to dynamically and automatically offer explanations that users can understand and act upon. This capability will make autonomy more trustworthy, useful, and transparent.

Incorporating object recognition processing is very important for providing explanations. For example, in the Atari game “Ms. Pacman”, the player controls Pac-Man through a maze, eating beans and avoiding monster enemies. There are also flashing dots known as power pellets that provide Pac-Man with the temporary ability to eat the enemies. By identifying the different objects (Pac-Man itself, beans, power pellets, enemies, walls, etc.), the deep RL agent can gain the potential to explain the actions like human beings.

In this paper, we develop a new method called \textbf{object saliency maps} to automatically produce object-level visual explanations that explain why an action was taken. The proposed method can be incorporated with any existing deep RL model to give human understandable explanation of why the model choose a certain action.

Our contributions are threefold: First, we propose a method to incorporate object characteristics to the learning process of deep reinforcement learning. ~Second, we propose a method to produce object-level visual explanation for deep RL models. ~ Third, our experiments show improvements over existing methods. 

\section{Related Work}
\subsection{Deep Reinforcement Learning}
Reinforcement learning is defined as learning a policy for an agent to interact with the unknown environment.~The rich representations given by deep neural network improves the efficiency of reinforcement learning (RL).~A variety of works thus investigate the application of deep learning on RL and propose a concept of deep reinforcement learning. Mnih et al. \cite{mnih2015humanlevel} proposed a deep Q-network (DQN) that combines Q-learning with a flexible deep neural network. DQN can reach human-level performance on many of Atari 2600 games but suffers substantial overestimation in some games \cite{DBLP:journals/corr/HasseltGS15}. Thus, a Double DQN (DDQN) was proposed by Hasselt et al. \cite{DBLP:journals/corr/HasseltGS15} to reduce overestimation by decoupling the target max operation into action selection and action evaluation. In the meantime, Wang et al. proposed a dueling network architecture (DuelingDQN) \cite{wang2015dueling} that decouples the state-action values into state values and action values to yield better approximation of the state value. 

Recent experiments of \cite{DBLP:journals/corr/MnihBMGLHSK16} show that the actor-critic (A3C) method surpasses the current state-of-the-art in the Atari game domain. Comparing to Q-learning, A3C is a policy-based model that learns a network action policy. However, for some games with many objects where different objects have different rewards, A3C does not perform very well. Therefore, Lample et al. \cite{lample2016playing} proposed a method that augments performance of reinforcement learning by exploiting game features. Accordingly, we propose a method of incorporating object features into current deep reinforcement learning models.

\subsection{Explainable Models}
There are some recent work on explaining the prediction result of black-box models. Erhan et al. \cite{erhan2009visualizing} visualised deep models by finding an input image which maximises the neuron activity of interest by carrying out an optimisation using gradient ascent in the image space. It was later employed by \cite{le2013building} to visualise the class models, captured by a deep unsupervised auto-encoder. Zeiler et al. \cite{zeiler2014visualizing} proposed the Deconvolutional Network (DeconvNet) architecture, which aims to approximately reconstruct the input of each layer from its output, to find evidence of predicting a class.  Recently, Simonyan et al. \cite{simonyan2013deep} proposed saliency maps to deduce the spatial support of a particular class in a given image based on the derivative of class score with respect to the input image. Ribeiro et al. \cite{ribeiro2016should} propose a method to explains the prediction of any classifier by local exploration, and apply it on image and text classification. All these models work at pixel level, and cannot explain the prediction at object level.

\subsection{Object recognition}
Object recognition aims to find and identify objects in an image or video sequence, where objects may vary in sizes and scales when translated or rotated. As a challenging task in the field of computer vision, Object Recognition (OR) has seen many approaches implemented over decades. 

Significant progress on this problem has been observed due to the introduction of low-level image features, such as Scale Invariant Feature Transformation (SIFT) \cite{lowe2004distinctive} and Histogram of Oriented Gradient (HOG) descriptors \cite{dalal2005histograms}, in sophisticated machine learning frameworks such as polynomial SVM \cite{mohan2001example} and its combination with Gaussian filters in a dynamic programming framework \cite{ronfard2002learning}. Recent development has also witnessed the successful application of selective search \cite{uijlings2013selective} on recognizing various objects. While HOG/SIFT representation can capture edge or gradient structure with easily controllable degree of invariance to local geometric and photometric transformations, it is generally acknowledged that progress slowed from 2010 onward, with small gains obtained by building ensemble systems and employing minor variants of successful methods \cite{mikolajczyk2004human}.

The burst of deep neural network over the past several years has dramatically improved object recognition capabilities such as convolutional neural network (CNN) \cite{lecun1989backpropagation} and other variants \cite{krizhevsky2012imagenet}, where the recent trend is to increase the number of layers and layer size \cite{sermanet2013overfeat}, while using dropout \cite{hinton2012improving} to address the problem of overfitting. Exploration of inception layers leads to a 22-layer deep model in the case of the GoogLeNet model \cite{szegedy2015going}. Furthermore, the Regions with Convolutional Neural Networks (R-CNN) method \cite{girshick2014rich} decomposes the overall recognition problem into two sub-problems and achieves the current state of the art performance. Recently, R-CNN has been optimized to reduce detection time and formulated as ``fast R-CNN'' \cite{girshick2015fast}, together with its various extensions \cite{karpathy2014deep} \cite{gupta2014learning} \cite{song2014learning}.

\section{Background}
\subsection{Reinforcement Learning Background}
\label{rl back}
Reinforcement learning tries to solve the sequential decision problems by learning from the history. Considering the standard RL setting where an agent interacts with an environment $\varepsilon$ over discrete time steps. In the time step $t$, the agent receives a state $s_t \in S $ and selects an action $a_t \in A$ according to its policy $\pi$, where $S$ and $A$ denote the sets of all possible states and actions respectively. After the action, the agent observes a scalar reward $r_t$ and receives the next state $s_{t+1}$. 

For example, in the Atari games domain, the agent receives an image input (consisting of pixels) as current state at time $t$, and chooses an action from the possible controls (Press the up/down/left/right/A/B button). After that, the agent receives a reward (how much the score goes up or down) and the next image input. 

The goal of the agent is to choose actions to maximize its rewards over time. In other words, the action selection implicitly considers the future rewards.  The discounted return is defined as  $R_t = \sum_{\tau =t}^\infty \gamma^{\tau-t} r_{\tau} $ where $\gamma \in [0,1]$ is a discount factor that trades-off the importance of recent and future rewards. 

For a stochastic policy $\pi$, the value of an action $a_t$ and the value of the states are defined as follows.
\begin{align}
Q^{\pi}(s_t, a_t) &= E [R_t|s=s_t, a=a_t, \pi] \\
V^{\pi}(s_t) &= E_{a\sim\pi(s_t)}[Q^{\pi}(s_t, a_t)]
\end{align}

The action value function (a.k.a., Q-function) can be computed recursively with dynamic programming: 
\begin{align}
Q^{\pi}(s_t, a_t) = E_{s_{t+1}} [r_t+\gamma E_{a_{t+1}\sim \pi(s_{t+1})}[Q^{\pi}(s_{t+1}, a_{t+1})]]
\end{align}

In value-based RL methods, the action value (a.k.a., Q-value) is commonly estimated by a function approximator, such as a deep neural network in DQN \cite{mnih2015humanlevel}. In DQN, let $Q(s, a;\theta )$ be the approximator parameterized by $\theta$. The parameter $\theta$ are learned by iteratively minimizing a sequence of loss functions, where the $i$th loss function is defined as:
\begin{align}
L_i(\theta_i) = E (r_t + \gamma \underset{a_{t+1}}{\max}Q(s_{t+1}, a_{t+1}; \theta_i) - Q(s_t, a_t; \theta_i))^2
\end{align}

In contrast to value-based methods, policy-based methods directly model the policy $\pi(a|s;\theta)$ and update the parameters $\theta$. For example, standard REINFORCE algorithm \cite{williams1992simple} updates the policy parameters $\theta$ in the direction $\bigtriangledown_\theta \log \pi(a_t|s_t;\theta)$. 

Actor-critic \cite{sutton1998reinforcement} architecture is a combination of value-based and policy-based methods. A learned value function $V^{\pi}(s_t)$ is considered as the baseline (the critic), and the quantity $A^{\pi}(a_t, s_t) = Q^{\pi}(a_t, s_t)-V^{\pi}(s_t)$ is defined as the \textit{advantage} of the action $a_t$ in state $s_t$, which is used to scale the policy gradient (the actor). Asynchronous deep neural network version of the actor-critic method (A3C) \cite{DBLP:journals/corr/MnihBMGLHSK16} gains state of the art performance in both Atari games and some other challenging domains.

We will enrich these models with ways to exploit object characteristics such as the presence and positions of objects in the training phase. Our envisioned architecture is shown in Figure.~\ref{fig:conv} and will be described in Section.~\ref{our methods}. 

\subsection{Object Recognition Background}
\label{obj back}
Object recognition is an essential research direction in computer vision. The common techniques include gradients, edges, linear binary patterns and Histogram of Oriented Gradients (HOG).~Based on these techniques, a variety of models are developed, including template matching, Viola-Jones algorithm and image segmentation with blob analysis \cite{MATLAB:2010}.~Considering our goal is to investigate whether object features can enhance the performance of deep reinforcement learning algorithms for Atari games, we use template matching to extract objects due to its implementation simplicity and good performance. 

Template matching, as a high-level computer vision technique, is used to locate a template image in a larger image. It requires two components -- source image and template image \cite{Brunelli:2009:TMT:1643435}. The source image is the one we expect to find matches to the template image while the template images is the patch image that is comparable to the template image. To identify the matching area, we slide through the source image with a patch (up to down, left to right) and calculate the patch's similarity to the template image.~In this paper, we use OpenCV \cite{itseez2015opencv} implement the template matching.~Specifically, we tried three different similarity measures methods. In the following equations, $T$ and $I$ denote two images while $x'$ and $y'$ are variable shift along x-direction and y-direction respectively. The $w$ and $h$ respectively represent width and height of the template image.
\begin{itemize}
	\item Square difference matching method\\
	Squared difference is a method that measures the pixel intensity differences between two images \cite{Ourselin2002}.~It computes the summation of squared product of two images' pixels subtraction.~Generally, this similarity measure directly uses the formulation of sum of squared error. It is chosen when speed matters. 
	\begin{align}
	R(x, y) = \sum_{x', y'}[T(x', y') - I(x+x', y+y')]^2
	\end{align}
	\item Correlation matching method\\
	The correlation matching method decides the matching point between source image and template image through searching the location with maximum value in the image matrices. It is chosen as similarity measure due to robustness \cite{cvpr2013Fast_Match}. However, it is computationally expensive compared with square difference matching method. 
	\begin{align}
	R(x, y) = \sum_{x', y'}[T(x', y')\cdot I(x+x', y+y')]^2
	\end{align}
	\item Correlation coefficient matching method\\
	Due to the limitation on speed of correlation matching method, correlation coefficient was proposed by \cite{cvpr2013Fast_Match} that performs transformation on both $T$ and $I$. The experimental results show it perform very well and is suitable for practical applications.
	\begin{align}
	&R(x, y) = \sum_{x', y'}[T(x', y')\cdot I'(x+x', y+y')]^2 \\
	&T'(x', y') = T'(x', y') - 1/(w \cdot h)\cdot \sum_{x'', y''} T(x'', y'')\\
	& I'(x+x', y+y') = I(x+x', y+y') - 1/(w\cdot h) \cdot \sum_{x'', y''}I(x+x'', y+y'')
	\end{align}
\end{itemize}
Considering the effect of changing intensity and template size, it is commonly applied the normalized version of the three methods above. Since accuracy is important for our task, we adopt normalized correlation coefficient matching method in template matching, which possesses the robustness of correlation matching method and time efficiency.
\section{Our Methods}
\label{our methods}
\subsection{Object-sensitive Deep Reinforcement Learning Model}
\label{O-DRL}
We propose an \textbf{Object-sensitive Deep Reinforcement Learning (O-DRL)} model that utilizes object characteristics (e.g., existence of a certain object and its position, etc.) in visual inputs to enhance the feature representation of deep reinforcement learning agents. The learned deep network would be more sensitive to the objects that appear in the input images, enabling the agent differentiates between ``good objects'' and ``bad objects''. Moreover, incorporating object recognition is very important for providing visual explanations, which would be further introduced in the Section.~\ref{object saliency}.

The key idea of the O-DRL model is to properly incorporate object features extracted from visual inputs to deep neural networks. In this paper, we propose to use \textbf{object channels} as a way to incorporate object features. 

The \textbf{object channels} are defined as follows: 
suppose we have $k$ objects detected in an image, we add $k$ additional channels to the original RGB channels of the original image. Each channel represents a single type of object. In each channel, for the pixels belong to the detected object, we assign value 1 in the corresponding position, and 0 otherwise. Through this, we successfully encode locations and categorical difference of various objects in an image.
\begin{figure}[t]
\centering
    \includegraphics[width=\textwidth]{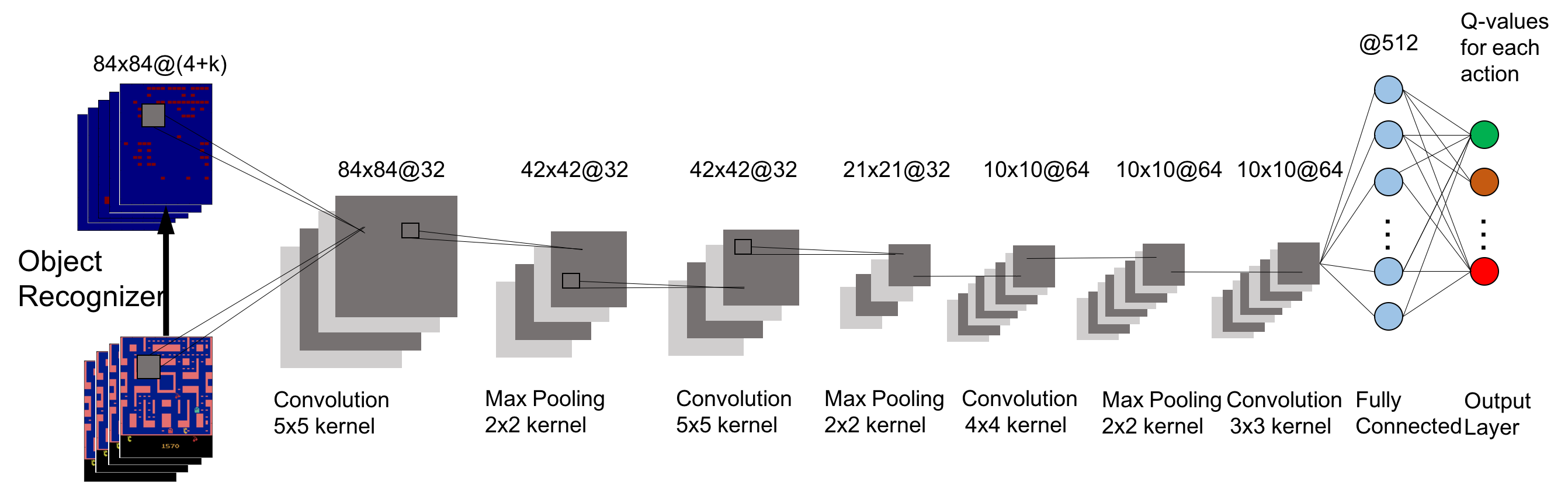}
    \caption{An example of neural network architecture for Object-sensitive Deep Q-network (O-DQN). Here, we get the screen image as input and pass it to the object recognizer to extract object channels. Then, the combined channels are given as input to the convolutional neural networks to predict Q-values.}\label{fig:conv}
\end{figure}
Take \textbf{Object-sensitive Deep Q-network (O-DQN)} as an example, the network architecture is shown in Figure~\ref{fig:conv}. Here, we get the screen image as input and pass it to the object recognizer to extract \textbf{object channels}. We also use a convolutional neural network (CNN) to extract image features just same as in DQN. Both object channels and the original state image are passed through the network to predict Q-values for each action. Note that this method can be adapted to different existing deep reinforcement learning frames, such as \textbf{Object-sensitive Double Q-Network (O-DDQN)} and \textbf{Object-sensitive Advanced Actor-critic model (O-A3C)}. In our experiments, all the object-sensitive DRL methods perform better than their non-object counterparts. 
\subsection{Object Saliency Maps}
\label{object saliency}
We propose a new method to produce \textbf{object saliency maps} for generating visual explanation of decisions made by deep RL agents. Before the introduction of \textbf{object saliency}, we first introduce \textbf{pixel saliency} \cite{simonyan2013deep}. This technique is first introduced to explain why a CNN classifies an image to a certain category. In order to generate explanation of why an DRL agent choose a certain action, we are interested in which pixels the model pays attention to when making a decision. For each state $s$, the model conduct action $a$ where $a = argmax_{a' \in A} Q(s, a')$. We would like to rank the pixels of $s$ based on their influence on $Q(s, a)$. Since the Q-values are approximated by a deep neural networks, the Q-value function $Q(s, a)$ is a highly non-linear function of $s$. However, given a state $s_0$, we can approximate $Q(s_0, a)$ with a linear function in the neighborhood of $s_0$ by computing the first-order Taylor expansion: 
\begin{align}
Q(s, a) \approx w^{T} s + b, 
\end{align}
where $w$ is the derivative of $Q(s,a)$ with respect to the state image $s$ at the point (state) $s_0$ and form the pixel saliency map:
\begin{align}
w = \frac{\partial Q(s,a)}{\partial s}|_{s_0}
\end{align}
Another interpretation of computing pixel saliency is that value of the derivative indicates which pixels need to be changed the least to affect the Q-value. 

\begin{figure}[H]
        \begin{subfigure}[b]{0.33\columnwidth}
                \centering
                \includegraphics[width=.85\linewidth]{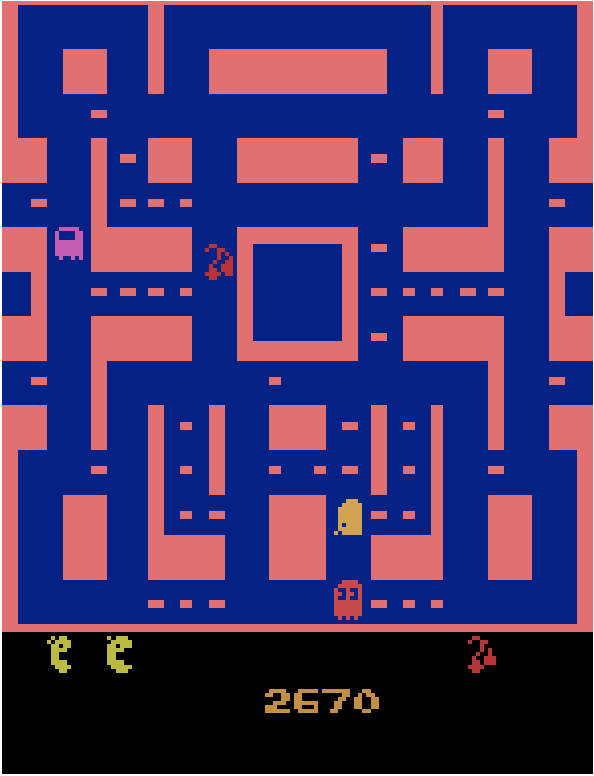}
                \caption{Original State}
                \label{ori}
        \end{subfigure}%
        \begin{subfigure}[b]{0.33\columnwidth}
                \centering
                \includegraphics[width=.85\linewidth]{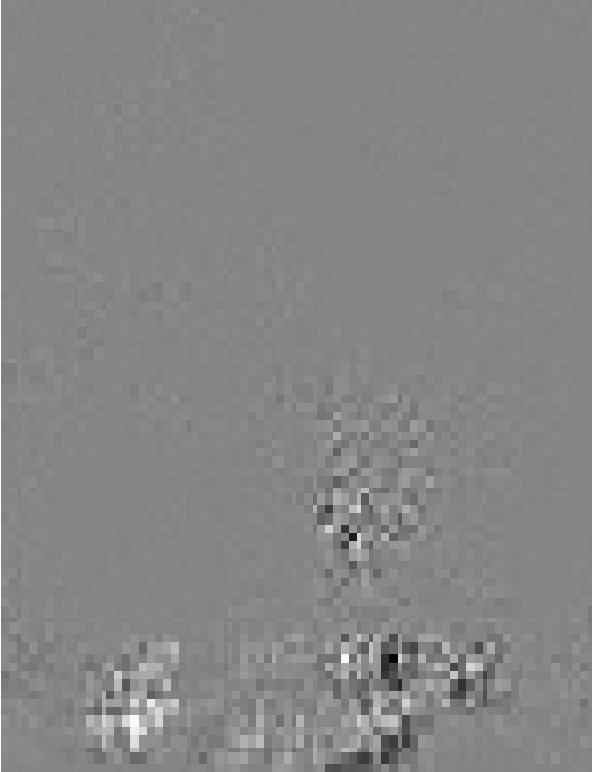}
                \caption{Pixel Saliency Map}
                \label{pixel}
        \end{subfigure}%
        \begin{subfigure}[b]{0.33\columnwidth}
                \centering
                \includegraphics[width=.85\linewidth]{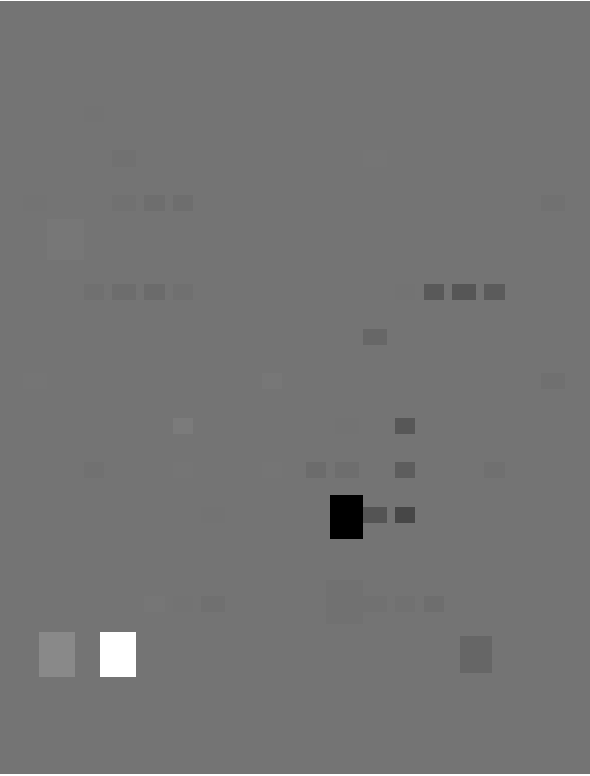}
                \caption{Object Saliency Map}
                \label{obj}
        \end{subfigure}%
        \caption{An example of original state, corresponding pixel saliency map and object saliency map produced by a double DQN agent in the game ``Ms. Pacman.''}\label{obj_exp}
\end{figure}

However, pixel-level representations are not obvious for people to understand. See Figure~\ref{obj_exp} for example. Figure~\ref{ori} is a screen image from the game Ms.Pacman. Figure~\ref{pixel} is the corresponding pixel saliency map produced by an agent trained with the Double DQN(DDQN) model. The agent chooses to go right in this situation. Although we can get some intuition of which area the deep RL agent is looking at to make the decision, it is not clear what objects the agent is looking at and why it chooses to move right in this situation.

To generate human understandable explanation, we need to focus on object level like humans do. Therefore, we need to rank the objects in a state $s$ based on their influence on $Q(s, a)$. However, it is nontrivial to design the derivative of $Q(s, a)$ with respect to the object area. Hence, we apply the following approach: for each object $O$ found in $s$, we mask the object with background color to form a new state $s_o$ as if the object does not appear in this new state. We calculate the Q-values for both states, and the difference of the Q-values actually represents the influence of this object on $Q(s, a)$. 
\begin{align}
 w = Q(s, a) - Q(s_o, a)
\end{align}

We can also derive that positive $w$ actually represents ``good'' object  which means the object gives positive future reward to the agent. And negative $w$ represents ``bad'' object since after we remove the object, the Q-value get improved. 

Figure~\ref{obj} shows an example of the object saliency map. While the pixel saliency map only explain a vague region where the model is looking at, the object saliency map can reveal which objects the model are looking at, and depicts the relative importance of each object. We can see that the object saliency map is more clear and meaningful than the pixel saliency map.

The computational cost of computing object saliency map is proportional to the number of objects we detected. If we have $k$ objects, the computational cost is $2k$ forward pass calculation of the model, which is affordable since $k$ is generally not too large, and one forward pass is fast in model testing time.
\section{Experiments}
\subsection{Baseline Methods}
We implemented deep-Q networks \textbf{(DQN)}\cite{mnih2015humanlevel}, double deep-Q networks \textbf{(DDQN)} \cite{DBLP:journals/corr/HasseltGS15}, dueling deep-Q networks \textbf{(Dueling)}\cite{wang2015dueling} and advanced actor-critic model \textbf{(A3C)}\cite{DBLP:journals/corr/MnihBMGLHSK16} described in Section~\ref{rl back} as baselines. We also implemented their object-sensitive counterparts by incorporating object channels described in Section~\ref{O-DRL}.
\subsection{Experiment Settings}
 We use OpenAI gym\cite{brockman2016openai} platform to perfoming our experiments. We choose 5 Atari 2600 games with distinguishable objects to test our models against baselines. They are Freeway, Riverraid, SpaceInvaders, BankHeist and Ms.Pacman.

We use the same network architecture for these DRL and O-DRL methods, shown in Figure~\ref{fig:conv}. The design is a little different from the original work of DQN \cite{mnih2015humanlevel} because of better performance achieved. There are four convolutional layers with 3 max pooling layers followed by 2 fully-connected layers. The first convolutional layer has 32 $5*5$ filters with stride 1, followed by a $2*2$ max pooling layer. The second convolutional layer has 32 $5*5$ filters with stride 1, followed by a $2*2$ max pooling layer. The third convolutional layer has 64 $4*4$ filters with stride 1, followed by a $2*2$ max pooling layer. The fourth and final convolutional layer has 64 $3*3$ filters with stride 1. The first full-connected layer has 512 hidden units. The final layer is the output layer, which differs in different models. In DQN, the dimension of the output layer is the number of actions. In A3C, two separate output layers are produced: a policy output layer with the dimension of the number of actions, a value output layer that contains only one unit.

We use 4 history frames to represent current state as described in \cite{mnih2015humanlevel}. For object representation, we use the last frame to extract object channels. In order to make objects distinct from each other, we do not use the reward clip strategy as described in \cite{mnih2015humanlevel}. Instead, we use the normalized rewards corresponding to the maximum reward received in the game. This is because the reward clip strategy assigns +1 for all rewards that are larger than 1 and -1 for all rewards that are smaller then -1, which makes different objects hard to distinguish. 
\subsection{Experiment Results}
\subsubsection{Object Recognition Results}
In order to verify the effectiveness of O-DRL methods, we need to verify the effectiveness of object recognition processing first. We adopt template matching with \textbf{correlation coefficient matching method} described in Section~\ref{obj back}. We manually create template images of objects in different games. For evaluation, we randomly select 100 game images from each of the game and label the objects in these images.

We use precision, recall and F1 score as our evaluation metrics for object recognition. Let $TP$, $FP$, $FN$ denote the number of true-positive, false-positive and false-negative predictions for all the labels, $Precision = \frac{TP}{TP+FP}, Recall = \frac{TP}{TP+FN}, F1 = 2 \frac{Precision \cdot Recall}{Precision + Recall}$. The number of object types in each game, as well as the precision, recall and F1 scores are reported in Table.~\ref{game stats}

\begin{table}[h]
\centering
\begin{tabular}{l|l|l|l|l|l}
                & Freeway & Riverraid & SpaceInvaders & BankHeist & Ms.Pacman \\\hline
\# Object types & 2       & 4         & 3             & 3         & 6        \\
Precision       & 1       & 1         & 1             & 1         & 1        \\
Recall          & 0.96    & 0.88      & 0.94          & 0.96      & 0.92     \\
F1  & 0.98    & 0.93      & 0.97          & 0.98      & 0.96    
\end{tabular}
\caption{Number of object types, precision, recall and F1 scores for 5 Atari 2600 games.}
\label{game stats}
\end{table}
We can see from the table that the precision of our object recognition is always 1, indicating the effectiveness of the template matching method. The F1 scores are also higher than 0.9. This indicates that the extraction of object channels is accurate and can be applied to object-sensitve DRL models.
\subsubsection{Object-senstive DRL Results}
We compare different DRL models with O-DRL models for 5 games. Each model is trained for 10 million frames. The average scores over 50 plays are presented in Table.~\ref{game results}. By comparing DRL models with their object-sensitive counterparts, we can see that object-sensitive DRL models perform better then DRL models in most of the games. All the best-performing models in each of the game are object-sensitive. We can also observe that O-DRL models only outperform DRL models by a small margin (1\%) in the game Freeway, but by a large margin (20\%) in the game Ms.Pacman. This may because that the game Ms.Pacman contain much more object types then the game Freeway, making object-sensitive models more effective.
\begin{table}[h]
\centering
\begin{tabular}{lllllllll}
              & DQN    & O-DQN  & DDQN    & O-DDQN  & Dueling & O-Dueling & A3C     & O-A3C   \\\hline
Freeway       & 26.9   & 27.1   & 29.3    & 29.6    & 21.5       & 21.8         & 30.3    & \textbf{30.7}    \\\hline
Riverraid     & 5418 & 5823 & 10324 & 10613 & 16983    & \textbf{18752}      & 12312 & 13215 \\\hline
SpaceInvaders & 1537 & 1528 & 2567  & 2585  & 5893     & 5923       & 23242 & \textbf{24027}   \\\hline
BankHeist     & 183  & 193  & 987   & 978   & 1056     & \textbf{1098}       & 956   & 1023  \\\hline
Ms.Pacman      & 1503   & 1593   & 2029    & \textbf{2453}    & 1614       & 1827         & 617     & 829    
\end{tabular}
\caption{Average scores over 50 plays of DRL and O-DRL models trained for 10 million frames on 5 Atari 2600 games. }
\label{game results}
\end{table}
\subsection{Case Study - Ms.Pacman}
We conduct a more detailed case study of the game Ms.Pacman to show the effectiveness of O-DRL models as well as the object saliency maps for explaining DRL models. We choose Ms.Pacman because it has most types of objects and it is hard to solve by DRL models. In this game, the player controls Pac-Man in a maze, collecting dots and avoiding ghosts. The actions contains left, right, down, up, leftup, leftdown, rightup, rightdown and nowhere, where ``nowhere'' represents do not press any button and continue the previous action.

\begin{figure}[H]
\centering
    \includegraphics[width=\textwidth]{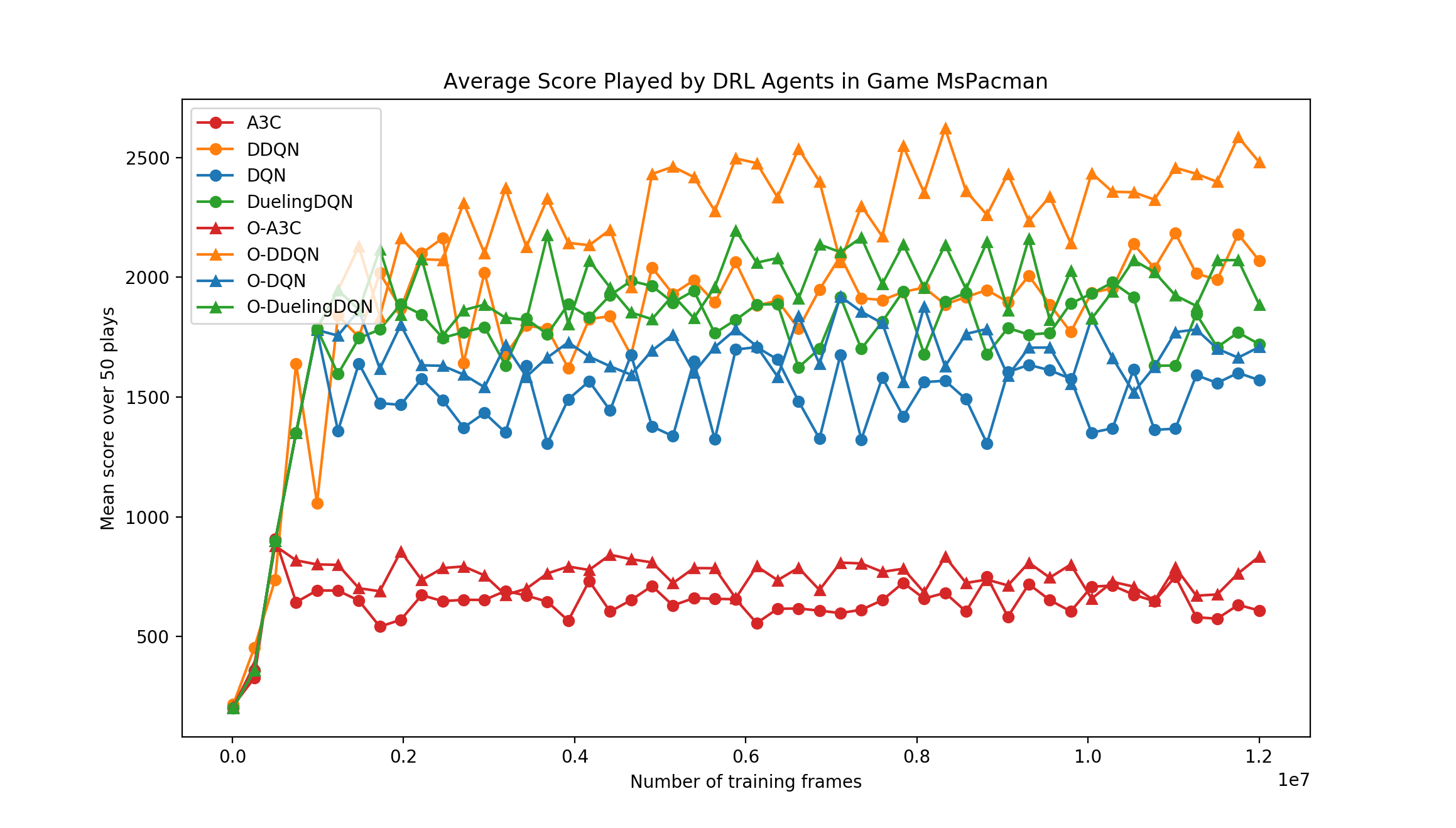}
    \caption{Average scores played by different DRL and O-DRL models in the game Ms.Pacman during the training process. The x-axis is the number of frames each model has been trianed for, y-axis is the average scores over 50 plays.}\label{fig:pacman}
\end{figure}

Figure.~\ref{fig:pacman} shows performance of different models during training. We can see that after about 2 million training frames, all the models converge to relative stable scores. We can also see that O-DRL models outperform their DRL counterparts by a non-trivial margin during the training phase. 
We also compare the object saliency maps and decisions made by the DRL model and the O-DRL model. We compare the DDQN model with the O-DDQN model in this case because they perform best in this game. We randomly sample 1000 states in a game play from human, and compare the decisions made by both models. We also produce object saliency maps for each model in order to see the which objects each model attend to when making decisions. 

Among 1000 samples, 98\% of the decisions given by both models are the same. Therefore, we look into the remaining 2\% of the samples.

\begin{figure}[H]
        \begin{subfigure}[b]{0.33\columnwidth}
                \centering
                \includegraphics[width=.85\linewidth]{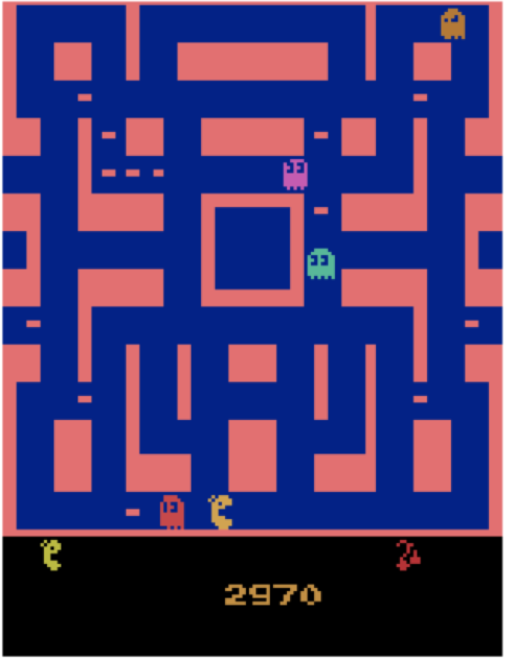}
                \caption{Current State}
                \label{cur1}
        \end{subfigure}%
        \begin{subfigure}[b]{0.33\columnwidth}
                \centering
                \includegraphics[width=.85\linewidth]{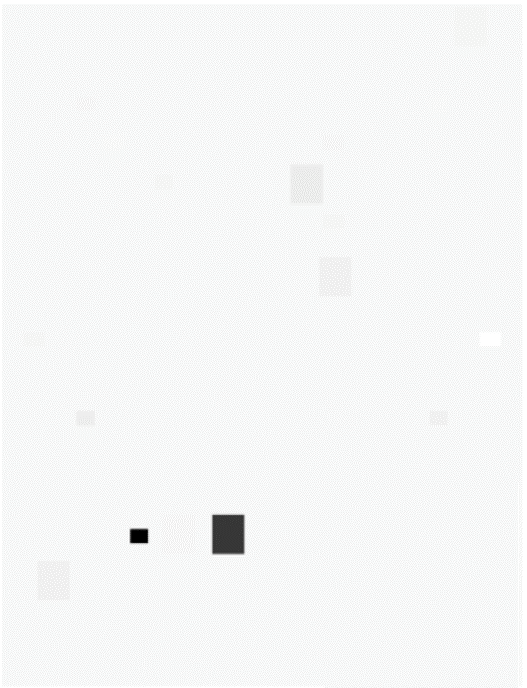}
                \caption{DDQN object saliency map}
                \label{DDQN1}
        \end{subfigure}%
        \begin{subfigure}[b]{0.33\columnwidth}
                \centering
                \includegraphics[width=.85\linewidth]{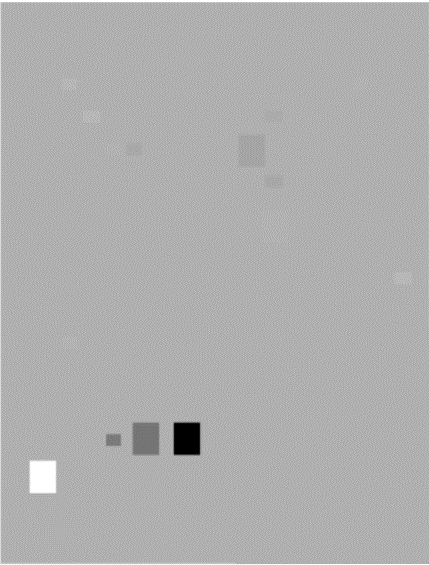}
                \caption{O-DDQN object saliency map}
                \label{O-DDQN1}
        \end{subfigure}%
        \caption{From left to right are the current state, the object saliency map produced by the DDQN model and the object saliency map produced by the O-DDQN model. The DDQN model chooses to go \textbf{left} while the O-DDQN model chooses to go \textbf{right} in this situation.}\label{obj-saliency-1}
\end{figure}

Figure.~\ref{obj-saliency-1} shows an example of different actions taken by both models and the object saliency maps produced by both models. In this state, Pac-Man is on the right of the ghost. In order to run away from the ghost, the human's choice is to go \textbf{right}. However, the DDQN model chooses to go \textbf{left}. We can see from Fig.~\ref{DDQN1} that the model does not focus on the ghost but the bean on the left of the ghost. Therefore, it chooses to go left without noticing of the ghost. The O-DDQN model successfully recognizes the ghost and choose to go \textbf{right}. We can also see this from Fig.~\ref{O-DDQN1}.

More detailed examples for how object saliency maps can help explaining the decisions made by each model can be found in the Appendix.~\ref{appendix:obj}. These prove that the object saliency maps can be used to visually explain why a model choose a certain action. 
\section{Conclusion and Future Work}
In this paper, we proved that by incorporating object features, we can improve the performance of deep reinforcement learning models by a non-trivial margin. We also proposed object saliency maps for visually explaining the actions taken by deep reinforcement learning agents. 

One interesting future direction is how to use object saliency maps to produce more human readable explanations like natural language explanations, for example, to automatically produce natural language explanations like ``I choose to go right to avoid the ghost''. 

Another direction is to test the ability of object features in a more realistic situation. For example, how to incorporate object features to improve the performance of auto-driving cars.

\section{Acknowledgement}
This research was supported by awards W911NF-13-1-0416 and FA9550-15-1-0442. Special thanks to Tian Tian and Jing Chen for insightful discussion and writing help.


\newpage
\label{sect:bib}
\bibliographystyle{plain}
\bibliography{easychair}

\begin{thebibliography}{10}

\bibitem{barto2004j}
MTRAG Barto.
\newblock J. 4 supervised actor-critic reinforcement learning.
\newblock {\em Handbook of learning and approximate dynamic programming},
  2:359, 2004.

\bibitem{brockman2016openai}
Greg Brockman, Vicki Cheung, Ludwig Pettersson, Jonas Schneider, John Schulman,
  Jie Tang, and Wojciech Zaremba.
\newblock Openai gym.
\newblock {\em arXiv preprint arXiv:1606.01540}, 2016.

\bibitem{Brunelli:2009:TMT:1643435}
Roberto Brunelli.
\newblock {\em Template Matching Techniques in Computer Vision: Theory and
  Practice}.
\newblock Wiley Publishing, 2009.

\bibitem{dalal2005histograms}
Navneet Dalal and Bill Triggs.
\newblock Histograms of oriented gradients for human detection.
\newblock In {\em 2005 IEEE Computer Society Conference on Computer Vision and
  Pattern Recognition (CVPR'05)}, volume~1, pages 886--893. IEEE, 2005.

\bibitem{erhan2009visualizing}
Dumitru Erhan, Yoshua Bengio, Aaron Courville, and Pascal Vincent.
\newblock Visualizing higher-layer features of a deep network.
\newblock {\em University of Montreal}, 1341:3, 2009.

\bibitem{girshick2015fast}
Ross Girshick.
\newblock Fast r-cnn.
\newblock In {\em Proceedings of the IEEE International Conference on Computer
  Vision}, pages 1440--1448, 2015.

\bibitem{girshick2014rich}
Ross Girshick, Jeff Donahue, Trevor Darrell, and Jitendra Malik.
\newblock Rich feature hierarchies for accurate object detection and semantic
  segmentation.
\newblock In {\em Proceedings of the IEEE conference on computer vision and
  pattern recognition}, pages 580--587, 2014.

\bibitem{gupta2014learning}
Saurabh Gupta, Ross Girshick, Pablo Arbel{\'a}ez, and Jitendra Malik.
\newblock Learning rich features from rgb-d images for object detection and
  segmentation.
\newblock In {\em European Conference on Computer Vision}, pages 345--360.
  Springer, 2014.

\bibitem{hinton2012improving}
Geoffrey~E Hinton, Nitish Srivastava, Alex Krizhevsky, Ilya Sutskever, and
  Ruslan~R Salakhutdinov.
\newblock Improving neural networks by preventing co-adaptation of feature
  detectors.
\newblock {\em arXiv preprint arXiv:1207.0580}, 2012.

\bibitem{itseez2015opencv}
Itseez.
\newblock Open source computer vision library.
\newblock {https://github.com/itseez/opencv}, 2015.

\bibitem{karpathy2014deep}
Andrej Karpathy, Armand Joulin, and Fei Fei~F Li.
\newblock Deep fragment embeddings for bidirectional image sentence mapping.
\newblock In {\em Advances in neural information processing systems}, pages
  1889--1897, 2014.

\bibitem{cvpr2013Fast_Match}
Simon Korman, Daniel Reichman, Gilad Tsur, and Shai Avidan.
\newblock Fast-match: Fast affine template matching.
\newblock In {\em Computer Vision and Pattern Recognition (CVPR), 2013 IEEE
  Conference on}, pages 1940--1947. IEEE, 2013.

\bibitem{krizhevsky2012imagenet}
Alex Krizhevsky, Ilya Sutskever, and Geoffrey~E Hinton.
\newblock Imagenet classification with deep convolutional neural networks.
\newblock In {\em Advances in neural information processing systems}, pages
  1097--1105, 2012.

\bibitem{lample2016playing}
Guillaume Lample and Devendra~Singh Chaplot.
\newblock Playing fps games with deep reinforcement learning.
\newblock {\em arXiv preprint arXiv:1609.05521}, 2016.

\bibitem{le2013building}
Quoc~V Le.
\newblock Building high-level features using large scale unsupervised learning.
\newblock In {\em Acoustics, Speech and Signal Processing (ICASSP), 2013 IEEE
  International Conference on}, pages 8595--8598. IEEE, 2013.

\bibitem{lecun1989backpropagation}
Yann LeCun, Bernhard Boser, John~S Denker, Donnie Henderson, Richard~E Howard,
  Wayne Hubbard, and Lawrence~D Jackel.
\newblock Backpropagation applied to handwritten zip code recognition.
\newblock {\em Neural computation}, 1(4):541--551, 1989.

\bibitem{lowe2004distinctive}
David~G Lowe.
\newblock Distinctive image features from scale-invariant keypoints.
\newblock {\em International journal of computer vision}, 60(2):91--110, 2004.

\bibitem{MATLAB:2010}
MATLAB.
\newblock {\em version 7.10.0 (R2010a)}.
\newblock The MathWorks Inc., Natick, Massachusetts, 2010.

\bibitem{mikolajczyk2004human}
Krystian Mikolajczyk, Cordelia Schmid, and Andrew Zisserman.
\newblock Human detection based on a probabilistic assembly of robust part
  detectors.
\newblock In {\em European Conference on Computer Vision}, pages 69--82.
  Springer, 2004.

\bibitem{DBLP:journals/corr/MnihBMGLHSK16}
Volodymyr Mnih, Adri{\`{a}}~Puigdom{\`{e}}nech Badia, Mehdi Mirza, Alex Graves,
  Timothy~P. Lillicrap, Tim Harley, David Silver, and Koray Kavukcuoglu.
\newblock Asynchronous methods for deep reinforcement learning.
\newblock {\em CoRR}, abs/1602.01783, 2016.

\bibitem{mnih2015humanlevel}
Volodymyr Mnih, Koray Kavukcuoglu, David Silver, Andrei~A. Rusu, Joel Veness,
  Marc~G. Bellemare, Alex Graves, Martin Riedmiller, Andreas~K. Fidjeland,
  Georg Ostrovski, Stig Petersen, Charles Beattie, Amir Sadik, Ioannis
  Antonoglou, Helen King, Dharshan Kumaran, Daan Wierstra, Shane Legg, and
  Demis Hassabis.
\newblock Human-level control through deep reinforcement learning.
\newblock {\em Nature}, 518(7540):529--533, February 2015.

\bibitem{mohan2001example}
Anuj Mohan, Constantine Papageorgiou, and Tomaso Poggio.
\newblock Example-based object detection in images by components.
\newblock {\em IEEE transactions on pattern analysis and machine intelligence},
  23(4):349--361, 2001.

\bibitem{Ourselin2002}
S{\'e}bastien Ourselin, Radu Stefanescu, and Xavier Pennec.
\newblock {\em Robust Registration of Multi-modal Images: Towards Real-Time
  Clinical Applications}, pages 140--147.
\newblock Springer Berlin Heidelberg, Berlin, Heidelberg, 2002.

\bibitem{ribeiro2016should}
Marco~Tulio Ribeiro, Sameer Singh, and Carlos Guestrin.
\newblock Why should i trust you?: Explaining the predictions of any
  classifier.
\newblock In {\em Proceedings of the 22nd ACM SIGKDD International Conference
  on Knowledge Discovery and Data Mining}, pages 1135--1144. ACM, 2016.

\bibitem{riedmiller2005neural}
Martin Riedmiller.
\newblock Neural fitted q iteration--first experiences with a data efficient
  neural reinforcement learning method.
\newblock In {\em European Conference on Machine Learning}, pages 317--328.
  Springer, 2005.

\bibitem{ronfard2002learning}
R{\'e}mi Ronfard, Cordelia Schmid, and Bill Triggs.
\newblock Learning to parse pictures of people.
\newblock In {\em European Conference on Computer Vision}, pages 700--714.
  Springer, 2002.

\bibitem{sermanet2013overfeat}
Pierre Sermanet, David Eigen, Xiang Zhang, Micha{\"e}l Mathieu, Rob Fergus, and
  Yann LeCun.
\newblock Overfeat: Integrated recognition, localization and detection using
  convolutional networks.
\newblock {\em arXiv preprint arXiv:1312.6229}, 2013.

\bibitem{simonyan2013deep}
Karen Simonyan, Andrea Vedaldi, and Andrew Zisserman.
\newblock Deep inside convolutional networks: Visualising image classification
  models and saliency maps.
\newblock {\em arXiv preprint arXiv:1312.6034}, 2013.

\bibitem{song2014learning}
Hyun~Oh Song, Ross~B Girshick, Stefanie Jegelka, Julien Mairal, Zaid Harchaoui,
  Trevor Darrell, et~al.
\newblock On learning to localize objects with minimal supervision.
\newblock In {\em ICML}, pages 1611--1619, 2014.

\bibitem{sutton1996generalization}
Richard~S Sutton.
\newblock Generalization in reinforcement learning: Successful examples using
  sparse coarse coding.
\newblock In {\em Advances in neural information processing systems}, pages
  1038--1044, 1996.

\bibitem{sutton1998reinforcement}
Richard~S Sutton and Andrew~G Barto.
\newblock {\em Reinforcement learning: An introduction}, volume~1.
\newblock MIT press Cambridge, 1998.

\bibitem{szegedy2015going}
Christian Szegedy, Wei Liu, Yangqing Jia, Pierre Sermanet, Scott Reed, Dragomir
  Anguelov, Dumitru Erhan, Vincent Vanhoucke, and Andrew Rabinovich.
\newblock Going deeper with convolutions.
\newblock In {\em Proceedings of the IEEE Conference on Computer Vision and
  Pattern Recognition}, pages 1--9, 2015.

\bibitem{uijlings2013selective}
Jasper~RR Uijlings, Koen~EA van~de Sande, Theo Gevers, and Arnold~WM Smeulders.
\newblock Selective search for object recognition.
\newblock {\em International journal of computer vision}, 104(2):154--171,
  2013.

\bibitem{DBLP:journals/corr/HasseltGS15}
Hado van Hasselt, Arthur Guez, and David Silver.
\newblock Deep reinforcement learning with double q-learning.
\newblock {\em CoRR}, abs/1509.06461, 2015.

\bibitem{wang2015dueling}
Ziyu Wang, Nando de~Freitas, and Marc Lanctot.
\newblock Dueling network architectures for deep reinforcement learning.
\newblock {\em arXiv preprint arXiv:1511.06581}, 2015.

\bibitem{williams1992simple}
Ronald~J Williams.
\newblock Simple statistical gradient-following algorithms for connectionist
  reinforcement learning.
\newblock {\em Machine learning}, 8(3-4):229--256, 1992.

\bibitem{zeiler2014visualizing}
Matthew~D Zeiler and Rob Fergus.
\newblock Visualizing and understanding convolutional networks.
\newblock In {\em European conference on computer vision}, pages 818--833.
  Springer, 2014.

\end{thebibliography}

\newpage
\appendix
\section{Examples of object saliency maps for the Game Ms.Pacman}
\label{appendix:obj}
\begin{figure}[H]
        \begin{subfigure}[b]{0.33\columnwidth}
                \centering
                \includegraphics[width=.85\linewidth]{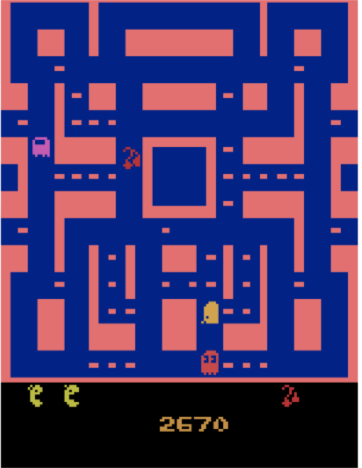}
                \caption{Current State}
                \label{exp3-ori}
        \end{subfigure}%
        \begin{subfigure}[b]{0.33\columnwidth}
                \centering
                \includegraphics[width=.85\linewidth]{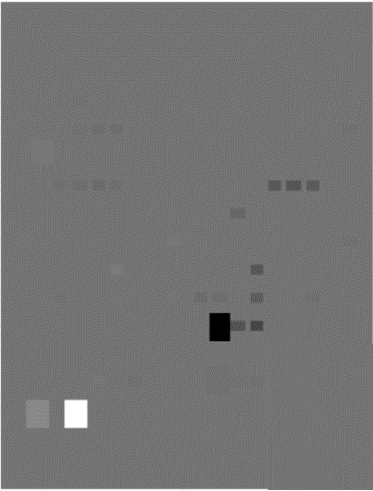}
                \caption{DDQN object saliency map}
                \label{exp3-DDQN}
        \end{subfigure}%
        \begin{subfigure}[b]{0.33\columnwidth}
                \centering
                \includegraphics[width=.85\linewidth]{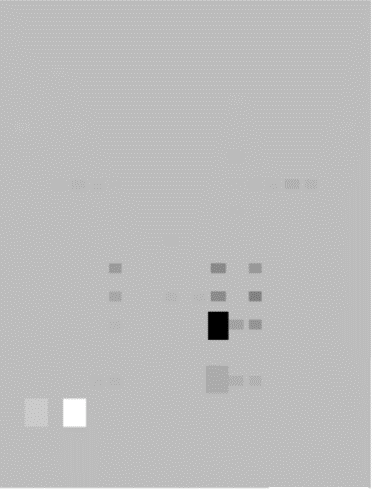}
                \caption{O-DDQN object saliency map}
                \label{exp3-O-DDQN}
        \end{subfigure}%
        \caption{From left to right are the current state, the object saliency map produced by the DDQN model and the object saliency map produced by the O-DDQN model. The DDQN model chooses to go \textbf{right} while the O-DDQN model chooses to go \textbf{up} in this situation.}\label{obj-saliency-3}
\end{figure}

\begin{figure}[H]
        \begin{subfigure}[b]{0.33\columnwidth}
                \centering
                \includegraphics[width=.85\linewidth]{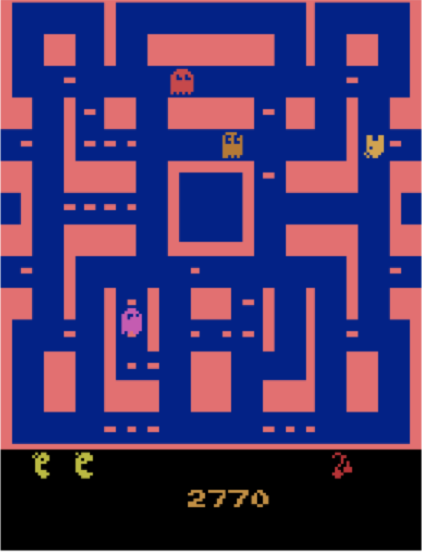}
                \caption{Current State}
                \label{exp2-ori}
        \end{subfigure}%
        \begin{subfigure}[b]{0.33\columnwidth}
                \centering
                \includegraphics[width=.85\linewidth]{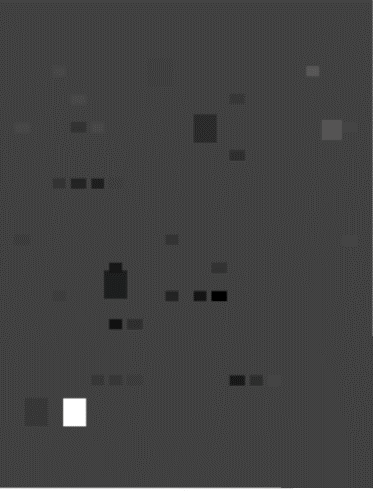}
                \caption{DDQN object saliency map}
                \label{exp2-DDQN}
        \end{subfigure}%
        \begin{subfigure}[b]{0.33\columnwidth}
                \centering
                \includegraphics[width=.85\linewidth]{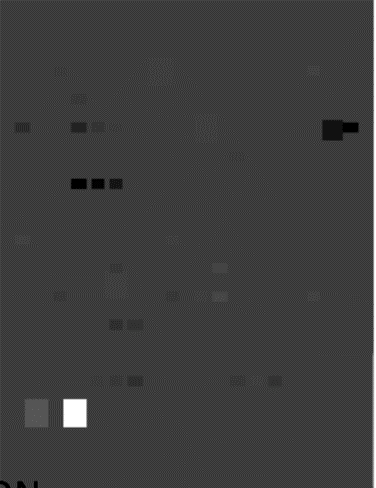}
                \caption{O-DDQN object saliency map}
                \label{exp2-O-DDQN}
        \end{subfigure}%
        \caption{From left to right are the current state, the object saliency map produced by the DDQN model and the object saliency map produced by the O-DDQN model. The DDQN model chooses to go \textbf{down} while the O-DDQN model chooses to go \textbf{right} in this situation.}\label{obj-saliency-2}
\end{figure}

Figure.~\ref{obj-saliency-3} shows another example of different actions taken by both models and the object saliency maps produced by both models. In this state, Pac-Man is on the upside of the ghost. The DDQN model chooses to go \textbf{right}. We can see from Fig.~\ref{exp3-DDQN} that the model focus on the beans on the right of Pac-Man. Therefore, it chooses to go right to eat the beans. The O-DDQN model chooses to go \textbf{up}. We can see from Fig.~\ref{exp3-O-DDQN} that the model concentrates on the beans on the upside of Pac-Man. Therefore it chooses to go up. Both of the decisions are reasonable becuase they both notice the ghost are on the downside of Pac-Man.
Figure.~\ref{obj-saliency-2} shows another example of different actions taken by both models and the object saliency maps produced by both models. In this state, Pac-Man is on the upper-right corner. The DDQN model chooses to go \textbf{down}. We can see from Fig.~\ref{exp2-DDQN} that the model focus on the ghosts and beans far from Pac-man. Therefore, it chooses to go down to eat the beans in the left-down side of Pac-Man. The O-DDQN model chooses to go \textbf{right}. We can see from Fig.~\ref{exp2-O-DDQN} that the model concentrates on the beans on the right of Pac-Man. Also according to the game setting, if the Pac-Man goes right, it will pass the right margin and appear in the left margin of the screen. The model successfully captures this rule and also focuses on the beans on the left margin of the screen.

\end{document}